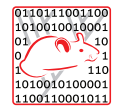

JOURNAL OF
BIOMEDICAL SEMANTICS

RESEARCH  Open Access

# Towards cross-lingual alerting for bursty epidemic events

Nigel Collier[1,2]

*From* Fourth International Symposium on Semantic Mining in Biomedicine (SMBM)
Hinxton, UK. 25-26 October 2010Correspondence: collier@nii.ac.jp
[1]National Institute of Informatics, 2-1-2 Hitotsubashi, Chiyoda-ku, Tokyo, Japan## Abstract

**Background:** Online news reports are increasingly becoming a source for event-based early warning systems that detect natural disasters. Harnessing the massive volume of information available from multilingual newswire presents as many challenges as opportunities due to the patterns of reporting complex spatio-temporal events.

**Results:** In this article we study the problem of utilising correlated event reports across languages. We track the evolution of 16 disease outbreaks using 5 temporal aberration detection algorithms on text-mined events classified according to disease and outbreak country. Using ProMED reports as a silver standard, comparative analysis of news data for 13 languages over a 129 day trial period showed improved sensitivity, F1 and timeliness across most models using cross-lingual events. We report a detailed case study analysis for Cholera in Angola 2010 which highlights the challenges faced in correlating news events with the silver standard.

**Conclusions:** The results show that automated health surveillance using multilingual text mining has the potential to turn low value news into high value alerts if informed choices are used to govern the selection of models and data sources. An implementation of the C2 alerting algorithm using multilingual news is available at the BioCaster portal http://born.nii.ac.jp/?page=globalroundup.## Introduction

As electronic data expands, online reports are coming to represent a new modality in early warning surveillance for natural disasters such as epidemics [1], typhoons and earthquakes [2,3]. Recent studies in disease surveillance such as [4] have shown that significant challenges still exist for fine-grained automated understanding of event dynamics.

Since 2006, BioCaster [5] has been performing gathering, semantic analysis and mapping of global news reports to provide a near-real time summary of human epidemics. The system is used regularly by both national and international health agencies as well as a growing base of individual users. Recent advances include expanding the number of diseases to include animal and crop pathogens as well as extending the number of languages from 4 to 13. With the increase in data came an understanding that public health analysts needed more help finding novel trends in the event stream.

© 2011 Collier; licensee BioMed Central Ltd. This is an open access article distributed under the terms of the Creative Commons Attribution License (http://creativecommons.org/licenses/by/2.0), which permits unrestricted use, distribution, and reproduction in any medium, provided the original work is properly cited.

In order to support the task of detecting the unusual, we compare five widely used temporal aberration detection algorithms to look for spikes in the news event stream. This paper builds on our previously reported study for monolingual news alerting [4] by seeking to explore the hypothesis that cross-lingual events from text mining can provide improved detection rates. Although we focus here on newswire as a source we believe the results should have applicability for other unverified reports such as email lists and the rapidly developing space of user generated content.

### Related work

The 2009 H1N1 pandemic illustrated how dependent each country is on the surveillance capacity in other states. Reducing public health risk depends on an overall strengthening of global health event monitoring as well as locally available sources such as clinical data and over-the-counter sales data. The Web provides a low cost surveillance infrastructure that has been shown to offer a timely means of detecting epidemics such as SARS [6] that is often several days ahead of the official reporting curve. In addition to work on BioCaster, there is a small but growing body of work looking at the issues of online public health monitoring such as GPHIN [6] and MedISys/PULS [7]. However, studies providing details of recall/precision/timeliness for end user tasks in media-based health surveillance are still surprisingly limited. To the best of our knowledge no previous study has explored the multilingual effects in this area. Several characteristics of early epidemic detection make the problem particularly challenging. Firstly, we want to catch epidemics as early as possible before they develop into humanitarian crises; Secondly, not every epidemic is of equal importance - those that are of most concern to the international community are described by the International Health Regulations [8]; Thirdly, patterns of media coverage are complex [9], at times focussing on dramatic and emotive imagery, at others prioritizing the reader's security and economic interests. In many ways the connection between media interest and the population at risk is often blurred.

How is this work different to various research in topic detection and tracking (TDT) [10] that has been undertaken for the last 14 years? Whilst both tasks look for events that are highly localized in time and space, the task we undertake begins with a predefined event semantics and a desire to distinguish the unexpected from the typical. Put another way, bursts in media interest do not always correspond to public health significance. The stream of work here seeks to uncover underlying trends and factors. Neither is this task entirely the same as TDT's first topic detection since we measure performance partly by the number of days before the silver standard that we can capture an event.

### Methods
#### Evaluation

In general it is extremely difficult to determine ground truth for the actual numbers and durations of disease outbreaks. As a silver standard we have chosen the best publicly available human network of reporters which is ProMED-mail [11]. ProMED-mail is a program of the International Society for Infectious Diseases with many expert volunteer reporters globally and a sophisticated staged editorial process. Outbreak reports are distributed to 40,000 subscribers by email, RSS feed and Web portal - precisely the audience we target in our automated system.

In this study we have used quite coarse-grained granularity by choosing countries and days as the units. This is due to the current limits of reliable location detection in the system and also the frequency of news that we observe. The recorded time for each event was normalized to system download time which takes place every hour of each day.

Evaluation uses the standard classification test measures of sensitivity (recall), specificity, positive predictive value (PPV or precision), negative predictive value (NPV) and timeliness. We also measured the average number of system alarms per 100 days and compared this to the silver standard. The F-measure (F1) is calculated in the usual way as the harmonic mean of sensitivity and PPV.

As in our previous study, the standard for a true positive was to obtain a system alert on a country-disease event on or before the silver standard alert. To allow for compatibility and comparison we kept the period for a qualifying system alert as up to 7 days prior to and including a qualifying ProMED report on the same topic. Other history period lengths might be more or less effective but were not the target of the investigation in this study. True positives were increased by 1 if there was any system alert that fell within the 7 day period. Multiple system alerts did not count twice. False positives were increased by 1 for each system alert that fell outside of the 7 day window. False negatives were counted as the number of qualifying alert periods when there were no system alerts. True negatives were counted as the number of days outside of any qualifying alert period when no system alert was given. In testing we tried to maximize F1 together with timeliness.

### Data

Figure 1 shows the 16 event streams that we explored. The events chosen for this study were determined based on diversity of geographical and media coverage rather than random selection. The 16 event streams contain 2064 surveillance days with 153 events (7.4% of alerting days) (Note that system data from the study will be made publicly available online for re-use via the GENI database interface on BioCaster). Since we wanted to explore the hypothesis that linguistic coverage in multiple languages could strengthen detection rates and timeliness we compared English news coverage against all languages including English for each of 16 disease outbreaks. English was chosen as the baseline because of its overall geographic representativeness. An alternative and perhaps more realistic approach might have been to use the native language for each outbreak country as the baseline which we will consider in future investigations. Because cross-lingual events on the 13 languages were only available in our system from December 2009, the trial period was from January to May 2010.

ProMED reports used in the silver standard excluded those that fell outside our case definition, based on the International Health Regulations [8] decision tree instrument. For example, requests for information, reports primarily focussed on control measures and aggregated summary reports not arising from specific events.

### Text mining system

The text mining system we explored involves a semantic pipeline of modules running on a high throughput cluster computer with 48 Xeon cores. Throughput is approximately 9000 articles per day. System news was gathered from multiple news sources through Google News and MeltWater News as well as specialized sources such as the European Media Monitor, IRIN and ReliefWeb. (Note that no ProMED-mail messages were included in the system data for this study using a block on the Internet domain and message title). In total this gives us access to over 80,000 news sources globally. The languages used in the study (in ISO-639-1) are: ar,zh,nl,en,fr,de,it, ko,pt,ru,es,vi and th.

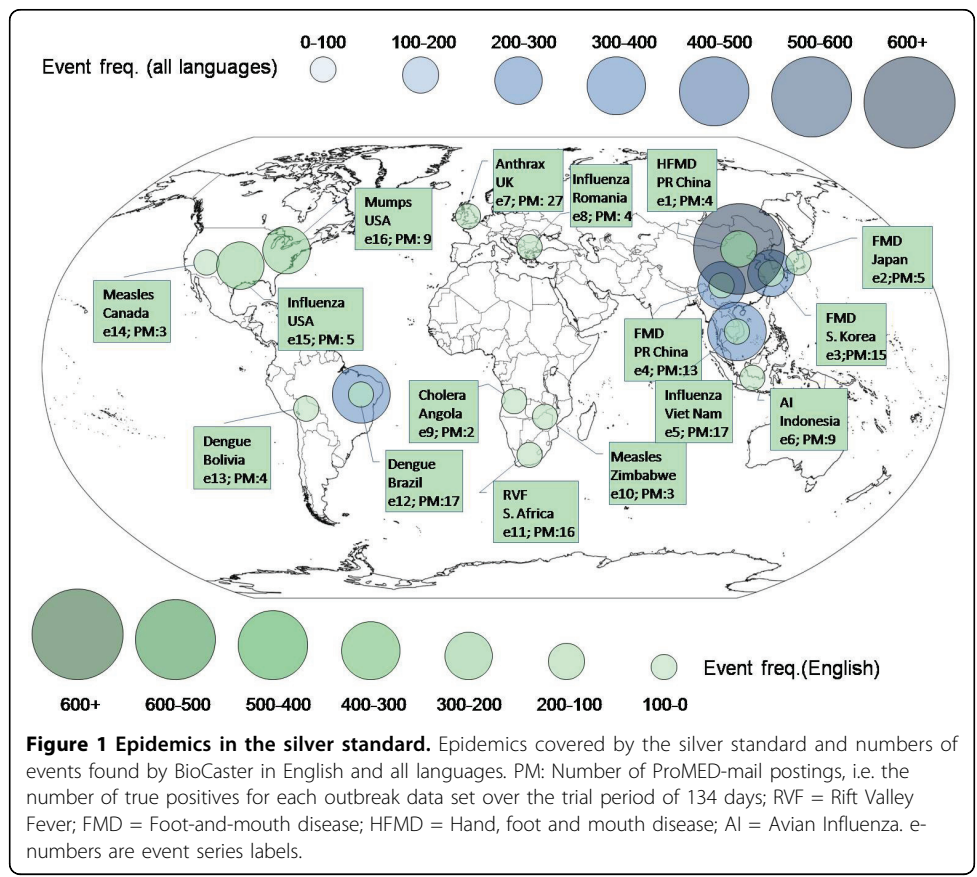

Figure 1 **Epidemics in the silver standard.** Epidemics covered by the silver standard and numbers of events found by BioCaster in English and all languages. PM: Number of ProMED-mail postings, i.e. the number of true positives for each outbreak data set over the trial period of 134 days; RVF = Rift Valley Fever; FMD = Foot-and-mouth disease; HFMD = Hand, foot and mouth disease; AI = Avian Influenza. e-numbers are event series labels.

Underlying the system is a publicly available multilingual application ontology [12] which is used within the rule books to make basic inferences such as countries from names of provinces, or diseases from causal pathogens. The BioCaster ontology (BCO) rules also allow us to unify variant forms of terms such as the 11 forms of A(H1N1).

After data sourcing, translation takes place from the twelve non-English languages used in this study using Google's online translation system. As a quality reference point we refer to a recent large-scale evaluation of machine translation for European language pairs [13]. In this study on news texts it was found that across a wide variety of metrics Google's online system consistently performed among the highest quality systems for Spanish-English, French-English and German-English language pairs.

Following machine translation, text classification using Naive Bayes (F1 0.93) removes non-disease outbreak news before text mining is applied. Rules are based on a regular expression matching toolkit called the Simple Rule Language [14] and divided between 18 entity types and template rules. The final structured event frames in XML includes slot values normalized to BCO root terms for disease, pathogen (virus or bacterium), time period, country and province. Additionally we also identify 15 aspects of public health events critical to risk assessment. For the purpose of this study we only made use of disease and country slots. Events in the 13 languages are treated in this study as being part of a univariate model for comparison purposes against English events.

Latitude and longitude of events down to the province level are found automatically using Google's API up to a limit of 15000 lookups per day, and then using lookup on 5000 country and province names harvested from Wikipedia.

### Alerting models

We experimented with a range of popular models for early alerting used in the public health community: the Early Aberration and Reporting System (EARS) quality control chart models C3, C2 and W2 as well as the F-statistic and the Exponential Weighted Moving Average (EWMA). All were implemented in Excel for the purpose of this study. The models are what might be termed 'snapshot' models because they all use short 7 day baselines that assume a relatively stationary background, i.e. ignoring medium to long term periodic variations such as seasonal cycles. The baselines are used to predict future trends against which the current day values are compared. All models also use a 2 day 'guard period' just before the target day $t$ to prevent the current day's data from being included in the baseline. All models use a minimally supervised method by setting a threshold parameter which we determined using the same 5 held out data sets used by [4]. These were 0.2 (C2 and W2), 0.3 (C3), 0.6 (F-statistic) and 2.0 (EWMA). A minimum standard deviation was set at 0.2 and a frequency purge was applied to remove singleton events, i.e. those with counts of 1 per day.

### C2

The EARS algorithms [15] are based on cumulative sum calculations commonly used in quality control. C2 triggers an alert when a test statistic $S_t$ exceeds a number $k$ of standard deviations above the baseline mean:

$$S_t = max(0, (C_t - (\mu_t + k\sigma_t))/\sigma_t) \tag{1}$$

where $C_t$ is the event count on the target day, $\mu_t$ and $\sigma_t$ are the mean and standard deviation of the counts during the baseline period. We set $k$ to 1 for all experiments.

### C3

C3 is a modified version of C2 so that the previous 2 observations (within the guard period) are added to the test statistic if the counts on those days does not exceed a threshold of 3 standard deviations plus the mean on those days. The rationale here is to extend the sensitivity of C2.

### W2

W2 [16] is a stratified version of C2 which compensates for weekend data outages by removing Saturday and Sunday data counts from the baseline. Alerting though can take place on any day.

### F-statistic

The calculation for the F-statistic [17] is:

$$S_t = \sigma_t^2 + \sigma_b^2 \tag{2}$$

where $\sigma_t^2$ approximates the variance during the testing window and $\sigma_b^2$ approximates the variance during the baseline window.

Calculation is as follows:

$$\sigma_t^2 = \frac{1}{n_t} \sum_{test}^{n_t} (C_t - \mu_b)^2 \qquad (3)$$

$$\sigma_b^2 = \frac{1}{n_b} \sum_{test}^{n_t} (C_t - \mu_b)^2 \qquad (4)$$

### EWMA

Unlike other models in our test, the EWMA provides for a non-uniformly weighted baseline by down-weighting counts that are on days further from the target day:

$$Y_1 = C_1 \qquad (5)$$

$$Y_t = \lambda C_t + (1 - \lambda) Y_{t-1} \qquad (6)$$

where $1 > \lambda > 0$ is a parameter that controls the degree of smoothing. The optimal level found from held out data was found to be 0.2. The test statistic is calculated as:

$$S_t = (Y_t - \mu_t) / [\sigma_t \times (\lambda / (2 - \lambda))^{0.5}] \qquad (7)$$

As above, $\mu_t$ and $\sigma_t$ are the mean and standard deviation on the baseline window.

### Results

Interestingly we found that approximately 80% of news reports covered only about half the ProMED-mail alert disease-country topics, implying that the remaining 20% of news has to provide coverage for almost half the topics. Surprisingly, the trend was broadly similar for both English and all language news. Although the sample size is relatively small, given that the events we chose were from all regions of the world, this implies that having news in more languages may have a deepening effect rather than a broadening effect on event coverage. The three notable exceptions were in the cases of FMD in China (e4 in Figure 1), Dengue in Brazil (e12) and Dengue in Bolivia (e13).

Results for global events on English (Table 1) show an advantage for the F-statistic if we are primarily concerned with sensitivity (recall) and alerting rates (shown in column [B]). However the F-statistic has a clear disadvantage with PPV (precision) which impacts heavily on the number of false alarms. This can be seen clearly by comparing the alarm rate per 100 days of 16.2 in column [A] with the ProMED average of 7.4. Both advantages and disadvantages are amplified when we add cross-lingual events.

Whilst the F-statistic has the highest overall F1, its high rate of false alarms reflected in the PPV makes it potentially an undesirable choice. If we seek for the best balance of F1 and timeliness with a minimum of false alarms then C3 looks like a more desirable alternative.

Cross-lingual event capture seemed to extend sensitivity in all models, improving F1 and timeliness. To see if we could harden our intuitions about these effects we looked

**Table 1 Aggregated results for global events**

| English only news (Worldwide) | | | | | | | |
|---|---|---|---|---|---|---|---|
| model | Se | Sp | PPV | NPV | Alarms[A] | Days[B] | F1 |
| C3 | 0.52 | 0.95 | 0.54 | 0.91 | 7.6 | 3.2 | 0.53 |
|  | (0.44,0.59) | (0.94,0.96) | (0.46,0.61) | (0.93,0.96) | | | |
| C2 | 0.42 | 0.97 | 0.57 | 0.92 | 5.1 | 3.1 | 0.48 |
|  | (0.34,0.50) | (0.96,0.98) | (0.48,0.66) | (0.93,0.96) | | | |
| W2 | 0.42 | 0.97 | 0.59 | 0.92 | 5.2 | 3.1 | 0.49 |
|  | (0.34,0.50) | (0.96,0.98) | (0.49,0.68) | (0.93,0.95) | | | |
| F-stat | 0.67 | 0.88 | 0.45 | 0.85 | 16.2 | 4.0 | 0.54 |
|  | (0.61,0.73) | (0.86,0.89) | (0.40,0.51) | (0.93,0.96) | | | |
| EWMA | 0.44 | 0.95 | 0.51 | 0.90 | 6.5 | 3.0 | 0.47 |
|  | (0.37,0.52) | (0.94,0.96) | (0.42,0.59) | (0.92,0.95) | | | |
| All language news (Worldwide) | | | | | | | |
| model | Se | Sp | PPV | NPV | Alarms[A] | Days[B] | F1 |
| C3 | 0.67 | 0.91 | 0.48 | 0.89 | 12.0 | 4 | 0.56 |
|  | (0.59,0.73) | (0.90,0.93) | (0.41,0.54) | (0.95,0.97) | | | |
| C2 | 0.54 | 0.95 | 0.49 | 0.91 | 7.1 | 3.7 | 0.51 |
|  | (0.46,061) | (0.94,0.96) | (0.42,0.56) | (0.95,0.97) | | | |
| W2 | 0.55 | 0.95 | 0.52 | 0.91 | 10.6 | 3.7 | 0.54 |
|  | (0.47,0.63) | (0.94,0.96) | (0.44,0.60) | (0.94,0.97) | | | |
| F-stat | 0.87 | 0.80 | 0.45 | 0.80 | 26.6 | 5.3 | 0.60 |
|  | (0.83,0.91) | (0.77,0.81) | (0.41,0.50) | (0.96,0.98) | | | |
| EWMA | 0.48 | 0.93 | 0.44 | 0.89 | 7.8 | 3.7 | 0.46 |
|  | (0.40,0.56) | (0.92,0.94) | (0.36,0.52) | (0.93,0.95) | | | |

Aggregated evaluation metrics for data sets e1 to e16 stratified by source language. The mean number of ProMED-mail alerts per 100 days was 7.4. [A] Model alarms per 100 days; [B] Mean number of days that alerts were given before ProMED-mail reports. Figures in parentheses show 95% CI.

specifically at South East Asia - a region where we would expect the representation of Chinese to be proportionately greater than English. Table 2 shows results which largely mirror those for the world as a whole. The noticeable exception though is that EWMA shows a large drop in performance.

## Discussion

Although the sample size is limited, the data suggests trends in model performance. C3 seems to perform best when we consider that the high false alarm rate for the F-statistic could desensitize users. Cross-language events generally seem to improve F1 performance by several points across most models except for EWMA. The benefits come from an extension in sensitivity but could be focussed on topics where we already have large coverage of English news. This is not to say that multilingual news is not useful, as we comment below, it could be that it has a greater role to play in extending detection rates of novel events at lower levels of geographic granularity than the country.

Beyond the cross-lingual effects, drill down analysis revealed that bag-of-words topic classification and event extraction using intra-sentential regular expressions were still letting through a proportion of non-events. We sampled 274 English news articles by hand from the BioCaster portal's implementation of the C2 algorithm using a 7 day baseline window and found that approximately 30% of positively classified news articles fell just below the borderline of our case definition. Commonly misclassified topics included: vaccination campaigns, factual advice on avoidance and treatment of

**Table 2 Aggregated results for SE Asian events**

| English only news (SE Asia) | | | | | | | |
|---|---|---|---|---|---|---|---|
| model | Se | Sp | PPV | NPV | Alarms[A] | Days[B] | F1 |
| C3 | 0.62 | 0.94 | 0.53 | 0.9 | 9.7 | 4.0 | 0.57 |
|  | (0.49,0.72) | (0.92,0.96) | (0.42,0.64) | (0.93,0.97) |  |  |  |
| C2 | 0.53 | 0.96 | 0.61 | 0.92 | 6.6 | 3.9 | 0.57 |
|  | (0.41,0.66) | (0.95,0.98) | (0.47,0.73) | (0.93,0.97) |  |  |  |
| W2 | 0.50 | 0.97 | 0.62 | 0.92 | 6.5 | 3.8 | 0.55 |
|  | (0.38,062) | (0.95,0.98) | (0.48,0.74) | (0.92,0.96) |  |  |  |
| F-stat | 0.76 | 0.83 | 0.42 | 0.82 | 20.9 | 5.0 | 0.54 |
|  | (0.67,0.84) | (0.80,0.86) | (0.35,0.50) | (0.94,0.97) |  |  |  |
| EWMA | 0.55 | 0.95 | 0.6 | 0.91 | 7.8 | 3.9 | 0.57 |
|  | (0.43,0.66) | (0.93,0.97) | (0.47,0.71) | (0.92,0.96) |  |  |  |
| All language news8 (SE Asia) | | | | | | | |
| model | Se | Sp | PPV | NPV | Alarms[A] | Days[B] | F1 |
| C3 | 0.71 | 0.91 | 0.50 | 0.89 | 13.4 | 4.9 | 0.59 |
|  | (0.60,0.80) | (0.88,0.93) | (0.41,0.59) | (0.94,0.97) |  |  |  |
| C2 | 0.62 | 0.94 | 0.50 | 0.91 | 8.3 | 4.3 | 0.55 |
|  | (0.48,0.74) | (0.92,0.96) | (0.38,0.62) | (0.94,0.98) |  |  |  |
| W2 | 0.61 | 0.94 | 0.53 | 0.91 | 17.1 | 4.6 | 0.57 |
|  | (0.49,0.73) | (0.92,0.96) | (0.41,0.65) | (0.94,0.97) |  |  |  |
| F-stat | 0.90 | 0.77 | 0.47 | 0.79 | 30.7 | 5.8 | 0.62 |
|  | (0.84,0.94) | (0.73,0.80) | (0.40,0.53) | (0.95,0.98) |  |  |  |
| EWMA | 0.53 | 0.94 | 0.48 | 0.89 | 8.1 | 3.9 | 0.50 |
|  | (0.40,0.65) | (0.91,0.96) | (0.36,0.61) | (0.92,0.96) |  |  |  |

Aggregated evaluation metrics for data sets e1 to e6 stratified by source language. The mean number of ProMED-mail alerts per 100 days was 8.1. [A] Model alarms per 100 days; [B] Mean number of days that alerts were given before ProMED-mail reports. Figures in parentheses show 95% CI.

infectious diseases, improvements in surveillance facilities and surveillance exercises. Often the articles mentioned an infectious disease and cited facts about case numbers such as "90% of cholera cases reported annually..." or an analysis of historical events. This points to a need to strengthen discourse analysis, such as inter-sentential causality and inclusion relations between events which the current intra-sentential template driven approach does not handle well.

The study also raises several questions about factors in the imbalance of reporting: why did Dengue in Brazil (e12) or FMD in China (e4) receive such massive local coverage but disproportionately less in the English media? Why did cholera in Angola (e9) or influenza in Romania (e8) receive comparatively low coverage overall? We also observed that the USA epidemics (e15 and e16) were widely reported in English but not so greatly in other languages.

In order to illustrate the potential complexity of the task we provide a detailed drill-down analysis of one of the outbreaks in our data set, i.e. cholera in Angola. Just to put the reporting of this outbreak into context: Angola itself is a former Portuguese colony which has suffered major outbreaks (e.g. 2006 to 2008) of cholera due to poor sanitation, drinking water infrastructure and environmental conditions. Although UNICEF has commented on recent advances, the country remains at risk, especially during the rainy season from January to mid-May.

21/1/2010 BioCaster detects 1 report in Spanish of 31 cases of cholera from October to December 2009. The report is republished in English and again in Spanish and Portuguese over the next few days. Since the report is for a historical outbreak (> 3 weeks old) it is a false positive.

**19/2/2010** ProMED-mail issues a report in English on cholera in Bocoio, Angola between 12/2/2010 and 18/2/2010. The cited source was the Angola Press Agency on 19/2/2010. BioCaster failed to capture this, so it is a false negative.

**4/3/2010** The Angop issues a report of 8 deaths from cholera in the province of Namibe. The report is cited by ProMED-mail on 19/3/2010.

**6/3/2010** BioCaster detects the 4/3/2010 report in its Spanish version. The status of this report is a false positive in the silver standard but should be considered as a true positive since it is a direct translation of a cited source used by ProMED-mail.

**10/3/2010** BioCaster detects reports in Spanish of a prevention campaign against cholera in Luanda. This seems to be a false positive but several such reports raise a system alarm. The high average number of reports means that smaller peaks of true positives on 14/3/2010 and 16/3/2010 do not raise system alarms. The F1 scores for multilingual reports are therefore lower than for English.

**14/3/2010** BioCaster detects 1 report in French from the Governor of Luanda requesting civil protection measures to prevent the proliferation of cholera following heavy rain and flooding. The indication of infrastructure stress is highly indicative of a true positive. Due to the high frequency of Spanish reports on 10/3/2010 no alarm is given.

**19/3/2010** ProMED-mail issues a report in English of cholera deaths in Tombua (Tombwa), southern Namibe, between 1/3/2010 and 3/3/2010. The cited source was Angop on 4/3/2010.

In this case BioCaster was more successful for English than for the multilingual system because a false spike of reports occluded subsequent true positives. In the case of the silver standard report on 19/3/2010, the cited English source was not detected but its Spanish translation was found a few days later - still much earlier than the ProMED-mail report.

The example is a relatively special case that illustrates an event that was not widely re-reported. The reports were made in English, Portuguese, French and Spanish from Angop. Externally, the 4/3/2010 article from Angop was republished in http://allafrica.com and http://africanseer.com on the 4th March. It was also referenced in a blog by the Namibia online community.

## Conclusions

Automated health surveillance using text mining is not intended as a substitute for skilled human analysts but as these results show, it does have the potential to reduce their information burden if informed choices are used to govern the selection of models. In order to help guide users in the significance of news events we implemented the C2 algorithm for multilingual news alerting in the 'Global Roundup' section of the BioCaster Web site at http://biocaster.nii.ac.jp. The results, updated each hour, show the test statistic value, the disease, country, province, focus species daily news frequency and the baseline mean and standard deviation. Additionally, citation links are provided to the news articles with a list up of all languages that contributed to the alert. The output is available as both RSS and a Twitter feed. Obvious improvements to the techniques described here could take place by modeling lower geographic granularity and reducing size differences between geo-units. More sophisticated approaches might incorporate proximity information between events or model how events propagate through news space. A more subtle effect of the granularity restriction is that the models we presented do not allow us to follow what might be called 'late warning' signals. i.e. follow on events within the country's borders. For this reason detecting events below the country level is desirable. Future work will need to concentrate on maximizing system sensitivity to overcome the fragmentation of the event distribution that occurs when we bucket events into smaller geographic units.

### List of abbreviations used
TDT: Topic Detection and Tracking; PPV: Positive Predictive Value; NPV: Negative Predictive Value; F1: F-score; BCO: BioCaster Ontology; EARS: Early Aberration and Reporting System; EWMA: Exponential Weighted Moving Average.

### Acknowledgements
I greatly acknowledge the comments by the anonymous SMBM 2010 reviewers. Funding support was provided in part by the Japan Science and Technology Agency under the PRESTO programme. The research work in its first unrevised form was presented at the SMBM 2010, Hinxton, Cambridge, U.K.
This article has been published as part of *Journal of Biomedical Semantics* Volume 2 Supplement 5, 2011: Proceedings of the Fourth International Symposium on Semantic Mining in Biomedicine (SMBM). The full contents of the supplement are available online at http://www.jbiomedsem.com/supplements/2/S5.

### Author details
[1]National Institute of Informatics, 2-1-2 Hitotsubashi, Chiyoda-ku,Tokyo, Japan. [2]Japan Science and Technology Agency, 2-1-2 Hitotsubashi, Chiyoda-kuTokyo, Japan.

### Author contributions
NC conceived, designed and carried out the evaluation.

### Competing interests
The author declares that he has no competing interests.

Published: 6 October 2011